\documentclass[preprint,12pt,numbers,authoryear]{elsarticle}

\usepackage{amssymb}
\usepackage{amsmath}
\usepackage{multirow}
\usepackage{makecell} 
\usepackage{booktabs}
\usepackage{tabularx}
\usepackage{adjustbox}
\usepackage{hyperref}
\usepackage{natbib}

\makeatletter
\long\def\@makecaption#1#2{%
  \vskip\abovecaptionskip\footnotesize
  \sbox\@tempboxa{#1 #2}%
  \ifdim \wd\@tempboxa >\hsize
    #1 #2\par
  \else
    \global \@minipagefalse
    \hb@xt@\hsize{\hfil\box\@tempboxa\hfil}%
  \fi
  \vskip\belowcaptionskip}
\makeatother

\makeatletter
\renewcommand{\figurename}{Fig.}
\long\def\fnum@figure{\figurename~\thefigure.}
\makeatother

\makeatletter
\long\def\fnum@table{\tablename~\thetable.}
\makeatother

\journal{Engineering Applications of Artificial Intelligence}

\begin{document}

\begin{frontmatter}



\title{Hoi2Threat: An Interpretable Threat Detection Method for Human Violence Scenarios Guided by Human-Object Interaction} 
\author[bit-me]{Yuhan Wang\fnref{equ}}
\author[bit-me]{Cheng Liu\fnref{equ}}
\author[bit-auto]{Daou Zhang}
\author[bit-me]{Zihan Zhao}
\author[bit-cs]{Jinyang Chen}
\author[bit-auto]{Purui Dong}
\author[bit-cs]{Zuyuan Yu}
\author[bit-ie]{Ziru Wang}
\author[bit-me]{Weichao Wu\corref{cor}}

\affiliation[bit-me]{organization={School of Mechatronical Engineering, Beijing Institute of Technology},
            postcode={100081},
            city={Beijing},
            country={China}}
\affiliation[bit-cs]{organization={School of Computer Scienece $\&$ Technology, Beijing Insitute of Technology},
            postcode={100081},
            city={Beijing},
            country={China}}
\affiliation[bit-auto]{organization={School of Automation, Beijing Insitute of Technology},
            postcode={100081},
            city={Beijing},
            country={China}}
\affiliation[bit-ie]{organization={School of Information and Electronics, Beijing Insitute of Technology},
            postcode={100081},
            city={Beijing},
            country={China}}

\cortext[cor]{Corresponding author. E-mail: wuweichao@bit.edu.cn}
\fntext[equ]{These authors contributed equally to this work.}

\begin{abstract}
In light of the mounting imperative for public security, the necessity for automated threat detection in high-risk scenarios is becoming increasingly pressing. However, existing methods generally suffer from the problems of uninterpretable inference and biased semantic understanding, which severely limits their reliability in practical deployment. In order to address the aforementioned challenges, this article proposes a threat detection method based on human-object interaction pairs (HOI-pairs), Hoi2Threat. This method is based on the fine-grained multimodal TD-Hoi dataset, enhancing the model’s semantic modeling ability for key entities and their behavioral interactions by using structured HOI tags to guide language generation. Furthermore, a set of metrics is designed for the evaluation of text response quality, with the objective of systematically measuring the model's representation accuracy and comprehensibility during threat interpretation. The experimental results have demonstrated that Hoi2Threat attains substantial enhancement in several threat detection tasks, particularly in the core metrics of Correctness of Information (CoI), Behavioral Mapping Accuracy (BMA), and Threat Detailed Orientation (TDO), which are 5.08, 5.04, and 4.76, and 7.10\%, 6.80\%, and 2.63\%, respectively, in comparison with the Gemma3 (4B). The aforementioned results provide comprehensive validation of the merits of this approach in the domains of semantic understanding, entity behavior mapping, and interpretability.
\end{abstract}

\begin{highlights}
\item The model uses structured tags for human-object interaction to guide text generation, enhance focus on key entities and behaviors, and reduce semantic comprehension bias.
\item A high-quality multimodal dataset for supervised training, which combines fine-grained annotation, effectively supports the training process to enhance the performance of the proposed method.
\item A set of systematic metrics for text response quality analysis is designed to quantify the model’s representation accuracy in interpreting threat events.
\item Adapt to the complex requirements of high-risk scenarios such as public safety protection and counter-terrorism and be deployed, providing users with highly reliable situational awareness results.
\end{highlights}


\end{frontmatter}



\section{Introduction}
\label{sec1}

In recent years, with increasing public security demand \citep{sultani2018real}, it has become imperative to accurately identify potential threats in daily life scenarios. Common violent scenes have encompassed a wide range of events, including violent street conflicts and terrorist attacks. The timely and accurate detection of potential threats in such scenes can provide law enforcement and security forces with early warning, thereby reducing casualties and property losses that would be that would be caused by violent events.

However, existing methodologies face many challenges when attempting to address violence. Conventional approaches predicated on manually designed features \citep{samaila2024video,leyva2017video,barragana2017unusual,singh2017graph,qin2018detecting}, including Scale Invariant Feature Transform (SIFT) \citep{Lowe2004SIFT}, Spatio-Temporal Interest Point (STIP) \citep{Laptev2003STIP}, and Histogram of Optical Flows (HOF) \citep{Laptev2008HOF}, predominantly depend on low-level image features for modeling. These methods have limited capacity to capture the complex semantics, scene dynamics, and interaction relationships embedded in violent behaviors, resulting in low detection accuracy in practical applications. Machine learning-based approaches \citep{yu2016content,ghasemi2017novel,wang2015abnormal,wen2015abnormal,cheng2016efficient} address the issue of inadequate feature representation to a certain degree by constructing discriminative models to enhance feature extraction and classification performance. However, such methods generally rely on large-scale feature engineering, which constrains the generalization ability of the models and makes it difficult to adapt to the development trends of scenario diversification. Deep learning-based approaches \citep{cheng2015video,ovhal2017analysis,sun2022detecting,wang2023memory,ramirez2021fall} automatically learn discriminative features through multilayer neural networks, significantly improving the performance of threat detection tasks. It is important to recognize that significant challenges persist in the implementation of these methods in high-risk environments. The fundamental issue pertains to the pervasive deficiency in interpretability and transparency within these models. These models predominantly depend on the potential statistical correlations present within the training data to formulate decisions, which hinders their ability to effectively address scenarios of a complex, volatile, and unforeseen nature within open environments.

In recent years, with the rapid development of visual-linguistic models (VLMs) \citep{touvron2023llama,chiang2023vicuna,jiang2023mistral,liu2024visual,bai2023qwen,dai2023instructblip,wang2023cogvlm}, multimodal information fusion has gradually become an important research direction in the field of threat detection. VLMs have achieved deep cross-modal semantic reasoning capability by jointly modeling visual and linguistic modalities, which has significantly enhanced the performance in complex semantic environments. In their seminal work,  \citet{zhang2024holmes} proposed a lightweight, time-based sampler that selects key frames with high-risk response from video frame sequences. This innovative approach effectively enhances the semantic expression ability of the model.  \citet{jiang2025lightweightdualbranchweaklysupervisedvideo} proposed an image-text dual-branching architecture, in which the implicit branch performs coarse-grained binary classification with the aid of visual features and extracts scene frames and actions. In contrast, the explicit branch employs the language-image alignment mechanism to perform fine-grained classification and introduces association rules in data mining as an auxiliary means of video description. This approach is intended to improve the performance of deeper semantic reasoning between modalities. The explicit branch utilizes linguistic and image alignment mechanisms for fine-grained classification and introduces association rules from data mining as an auxiliary means of video description to achieve interpretability of threat monitoring. However, the aforementioned methods rely exclusively on coarse-grained explicit matching between visual features and textual descriptions, thus lacking semantic modeling of key entities, behaviors, and their interactions in images. This deficiency results in biased semantic understanding.

\begin{figure}[htbp]
    \centering
    \includegraphics[width=\textwidth]{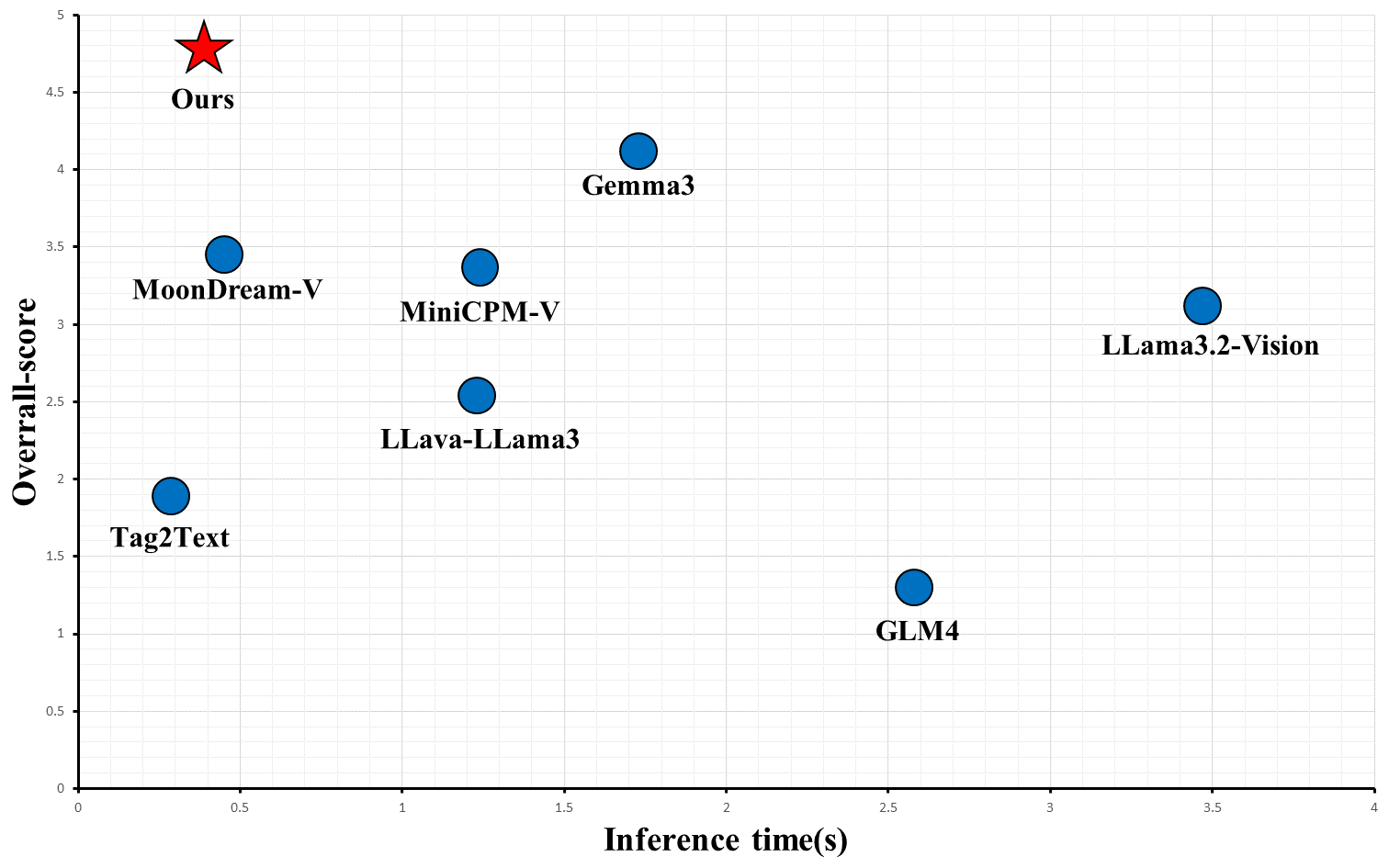}
    \caption{A thorough examination of the performance of existing LLM methods is conducted on a variety of datasets. The Overall-score is the weighted average score of different indicators in Table \ref{tab1}. below. This score takes different evaluation indicators into account and can comprehensively reflect the performance of threat detection methods. The blue circles in the figure represent other threat detection methods, while the red stars represent the method under investigation.}
    \label{fig1}
\end{figure}

In order to address the challenges posed by the interpretability of inference and the presence of semantic understanding bias in existing research, this article presents a multimodal threat detection method, Hoi2Threat. The proposed approach involves the construction of a high-quality multimodal dataset, TD-Hoi, which integrates fine-grained image annotations, including human-object interactions (HOI) and textual descriptions. This integration serves to enhance the interpretability of the model in performing high-risk tasks, thereby ensuring enhanced decision reliability. On this basis, we design structured tags based on HOI to guide text generation and strengthen the model's focus on key entities and behaviors, with a view to reducing semantic comprehension bias. The experimental results demonstrate the capacity of Hoi2Threat to accurately establish the mapping relationship between entities and actions. The existing methods are assessed in terms of threat event identification, entity-action mapping, and interpretability in a wide range of datasets. The proposed method demonstrates strong adaptability and deployment reliability in high-risk scenarios, such as threat detection. The specific results are shown in Fig. \ref{fig1}.

The following list enumerates the primary contributions of the present article:
\begin{itemize}
\item A multimodal dataset, TD-Hoi, integrates images and fine-grained annotations, including human-object interactions (HOI) and textual descriptions. The integration of these components is intended to enhance the interpretability of inferences and the reliability of decisions. Additionally, it is expected to provide a secure foundation for applications in human violence scenarios.
\item A multimodal threat detection method based on HOI structuring is proposed. The method has been demonstrated to establish the mapping relationship between threat entities and behaviors in complex threat scenarios, and to realize the accurate discrimination of threats.
\item A set of systematic metrics for analyzing text response quality has been designed. The metrics are capable of quantifying the representation accuracy of the model in the interpretation of image threat events, thereby effectively addressing the current lacuna in the field of multimodal semantic response quality assessment.
\end{itemize}

The remainder of this article is organized as follows. Section \ref{sec2} details the dataset TD-Hoi we built, covering data collection and processing. Section \ref{sec3} presents the model architecture, analyzing each module of the proposed Hoi2Threat and elaborating on the selection and design of the training loss function. Section \ref{sec4} shows the experimental results, including a comparison with existing methods and ablation experiments to evaluate our approach. Section \ref{sec5} concludes this article, summarizing the main work and the findings.

\section{Data benckmark}
\label{sec2}
\begin{figure}[t]
\centering
\includegraphics[width=\textwidth]{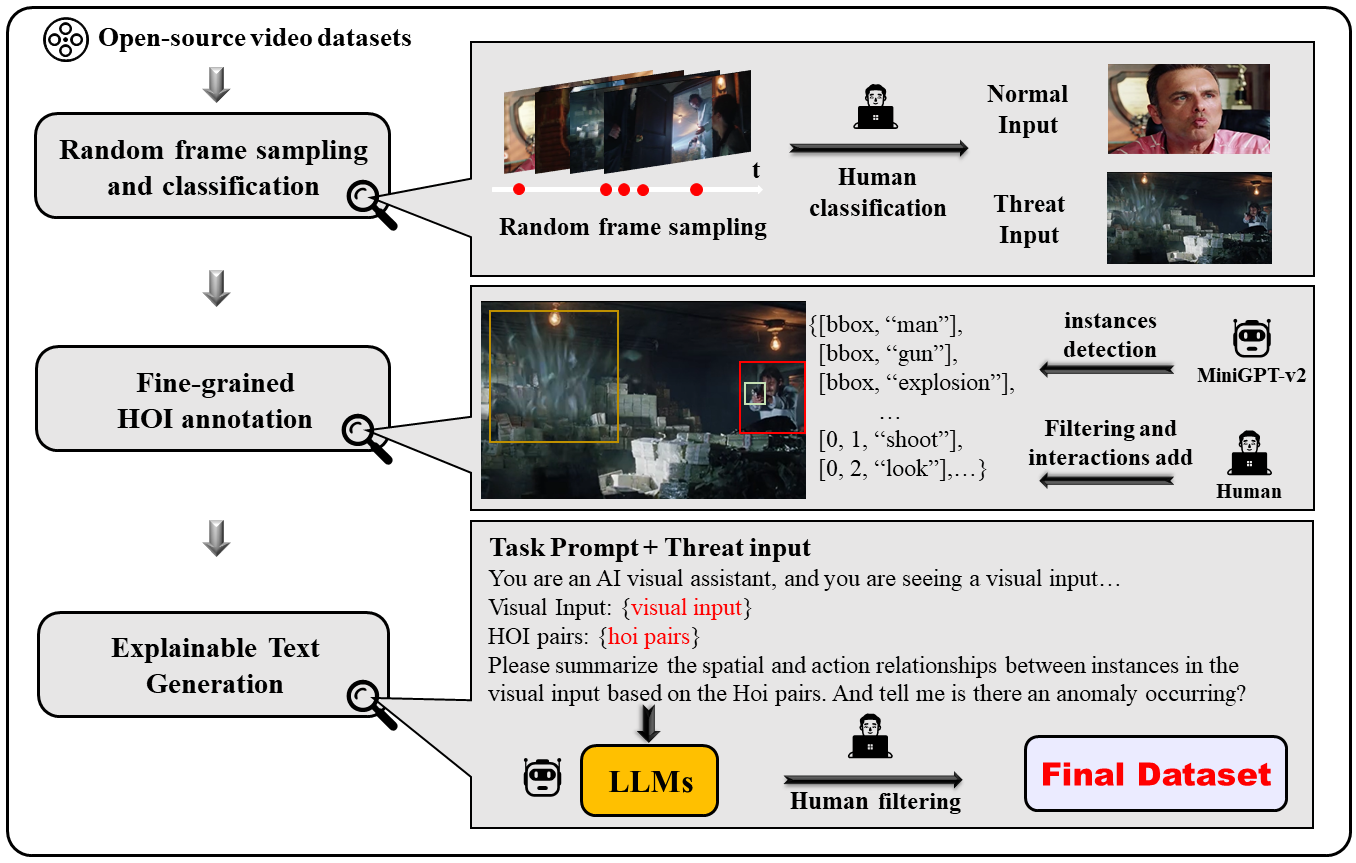}
\caption{TD-Hoi data engine. Initially, a collection of images is obtained from open-source datasets. Secondly, we employ MiniGPT-v2  for fine-grained annotation and incorporate behavioral mapping between disparate entities. Finally, we employ GPT-4.1 to construct interpretable text data. Throughout the annotation process, the efficiency and quality of data annotation are ensured by close cooperation between manual and LLMs.}\label{fig2}
\end{figure}

Threat events typically exhibit a multifaceted characterization, comprising intertwined threat behaviors and related entities \citep{nayak2021comprehensive}. Among them, threat behavior refers to dynamic processes with threat semantics, including but not limited to assaults, shootings, hijackings, and other violent acts. Relevant entities are defined as individuals or objects with offensive properties that are highly related to threatening behaviors. These entities may include terrorists, victims, and various types of violent tools such as firearms and knives. The manifestation of these threatening behaviors and related entities frequently serves as a pivotal indicator of the occurrence of a threatening event. Therefore, a multimodal dataset, TD-Hoi, is constructed, incorporating fine-grained mapping relationships between humans, objects, and actions, as well as description texts. The overall flow of the data annotation engine is illustrated in Fig. \ref{fig2}.

In this section, the construction process of this dataset is systematically introduced, covering raw data acquisition, fine-grained labeling strategy, and interpretable text generation mechanism. The TD-Hoi data engine employs multimodal information fusion to provide high-quality, semantically rich, and interpretable data support for threat detection tasks. 
\subsection{Data collection}
\label{subsec2.1}

The visual data presented herein is collected from the training set of XD-Violence \citep{wu2020not}, a dataset that is identified as one of the most extensive threat detection datasets. A significant advantage of XD-Violence over UCF-Crime \citep{sun2024multimodal} is its superior image quality, which is complemented by a comprehensive array of data sources and comprehensive supervisory information. To ensure the balance of the dataset, it is supplemented with normal images exhibiting HOI selected from the most extensive generalized dataset, COCO \citep{lin2014microsoft}. A total of 8,282 images were collected through a combination of random frame sampling and manual classification, including 3,311 images depicting threats and 4,971 images classified as normal.

\subsection{Fine-grained HOI annotation}
\label{subsec2.2}

This subsection is designated for the labeling of the collected images with HOI-pairs, thereby ensuring the provision of fine-grained semantic information for the model. Specifically, we employ MiniGPT-v2 to detect and localize entities in the images, with an average of 2.70 entities labeled per image. Consequently, the detected entities $\mathcal{E}=\left\{e_i\right\}^{N_{obj}}$ are associated with non-threatening or threatening behavioral relationships, as illustrated in Eq. (\ref{eq:1}):

\begin{equation}
\begin{aligned}
\mathcal{H} = \left\{img_n, [e_i, e_j, act_{ij}], \cdots, [e_k, e_l, act_{kl}] \right\}^{N_e}
\quad i, j, k, l \in N_e.
\end{aligned}
\label{eq:1}
\end{equation}

\noindent
where $\mathcal{H}$ denotes the set of processed HOI-pairs, $\textit{img}_n$ denotes the nth image that is labeled, $N_e$ expresses the total number of entities identified in the image, $e_i$ is the $i^{\text{th}}$ entity detected in $\textit{img}_n$, and $\textit{act}_{ij}$ denotes the behavioral relationship between the $i^{\text{th}}$ and the  $j^{\text{th}}$  entity. Ultimately, we use $\mathcal{H}$ to train the HOI encoder individually so that it can effectively learn the association between HOI and threat detection.

\subsection{Training texts generation based on HOI-pairs}
\label{subsec2.3}

The primary objective of the text generation training phase is to ensure precise alignment of the text with the HOI-pairs, thereby guaranteeing that the generated content provides comprehensive coverage of each set of HOI elements. This alignment strategy effectively enhances the model's understanding of HOI, thus improving its performance in related tasks. First, GPT-4.1 \citep{OpenAI_GPT41_2024} and the predefined HOI-pairs are utilized to discriminate the scenes in the images and generate the explanatory text, $c_n$, that covers the interaction elements and semantic logic. Second, the text generation process is based on manually designed prompts to ensure that each interaction element is covered and the relationship structure is accurately presented, thereby enhancing the model's comprehension of human-object interactions. To ensure the highest standards of textual quality, the generated text undergoes a meticulous review process that encompasses the evaluation of accuracy, coherence, and clarity of expression. The integration of text responses and HOI-pairs culminates in the formation of a high-quality dataset, which serves as a foundational element for subsequent model training.

\section{Hoi2Threat}
\label{sec3}
Existing VLM research lacks an explicit semantic decomposition and inference mechanism in the Image-text modeling process. Consequently, research frequently generates text directly under the premise of not yet fully understanding the semantics and details of the image. This is susceptible to machine hallucinations. Notably, through the key information transfer method of image-tag-text, the model can explicitly receive intermediate semantic guidance during the generation process. This enhances the semantic consistency of text generation and, in turn, strengthens the model’s interpretability in complex tasks.

\begin{figure}[t]
\centering
\includegraphics[width=\textwidth]{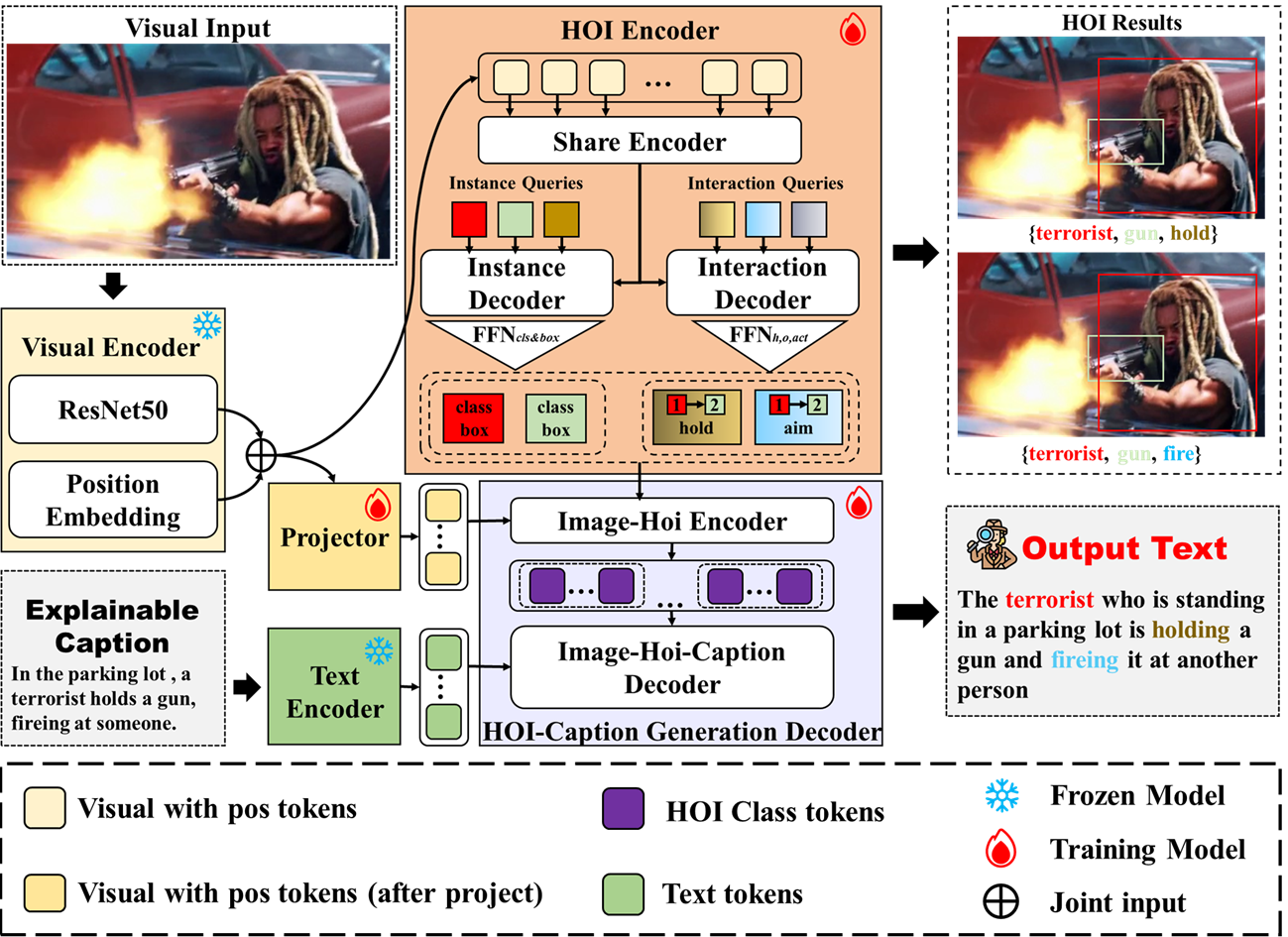}
\caption{Overview of the Hoi2Threat framework. The approach utilizes a visual image as the input source and generates a set of interpretable text in response to a threat event. The HOI encoder first extracts HOI-pairs from the image, which are then integrated with the original visual features through a cross-attention mechanism. These integrated features are subsequently fed into an Image-HOI-Caption decoder, which generates a semantically complete textual response.}\label{fig3}
\end{figure}

Based on these findings, this article puts forth a tag-guided multimodal approach, Hoi2Threat, which is predicated on the image-tag-text information transfer mechanism. The approach utilizes HOI-pairs as the intermediate semantic tags to enhance the semantic alignment and relational modeling capabilities during the generation process. The use of HOI-pairs is demonstrated to be an effective means of capturing the key entities and their behavioral relationships in the violence scenario through fine-grained standardized information. As shown in Fig. \ref{fig3}, the general structure of Hoi2Threat is presented..

\subsection{Model architecture}

\subsubsection{HOI encoder}
\label{subsubsec3.1.1}

In this article, HOTR \citep{kim2021hotr} is adopted as the HOI encoder $\phi_{hoi}$. It extracts visual features of the image using a shared CNN \citep{carion2020end}. Then, it models instances and behaviors through a parallel entity decoder and a behavior decoder to obtain entity and behavioral representations. The specific flow is illustrated in Fig. \ref{fig3}. The HOI encoder enhances the modeling capability of HOI by means of the HOI pointer mechanism. This mechanism allows for accurate correlation of behaviors with corresponding entities. The behavioral decoder acquires the behavioral representation, denoted by $z_i$, and processes it through two feed-forward networks, $\text{FFN}_h:\mathbb{R}^d\to\mathbb{R}^d$ and $\text{FFN}_o:\mathbb{R}^d\to\mathbb{R}^d$ to point to the associated human or object entity vectors $v_i^h$ and $v_i^o$. These vectors are defined as follows: $v_i^h=\text{FFN}_h(z_i)$ and $v_i^o= \text{FFN}_o(z_i)$. Finally, the human or object HOI pointers (that is, ${c}_i^h$ and ${c}_i^o$) are obtained by Eq. (\ref{eq:2}).

\begin{equation}
    \begin{aligned}
        &\hat{c}_i^h = \arg\max_j \text{sim}(v_i^h, \mu_j) \\
        &\hat{c}_i^o = \arg\max_j \text{sim}(v_i^o, \mu_j),
    \end{aligned}
    \label{eq:2}
\end{equation}

\noindent
where $\mu_j$ is the  $j^{\text{th}}$  entity representation and  $\text{sim}=u^Tv/||u||||v||$.

Given the $\gamma$ behavioral categories, two feedforward networks, denoted as ${\rm FFN}_{box}:\mathbb{R}^d\rightarrow\mathbb{R}^4$ and ${\rm FFN}_{act}:\mathbb{R}^d\rightarrow\mathbb{R}^\gamma$, are employed to facilitate the recombination of HOI pointers, thereby yielding the accurate HOI triples. The final HOI prediction is obtained after processing by these two networks, as demonstrated in Eq. (\ref{eq:3}):

\begin{equation}
    {Hoi}_i=\left\{\left[b_i^h,b_i^o,h_i,o_i,a_i\right]\right\}_{i=1}^K,
    \label{eq:3}
\end{equation}

\noindent
where $h_i$, $o_i$, $a_i$ are the human entities, object entities and their behavioral mappings detected by ${Image}_i$ by HOI encoder, $b_i^h$, $b_i^o$ represent the positions of human entities and object entities in the image, respectively. $Hoi$ is the collection of real HOI triples. $K$ denotes that HOI encoder predicts a total of K HOI triples, each of which consists of a human frame, an object frame, and a behavioral binary classification. Each prediction captures a unique person pair and may contain multiple interactions. The loss of HOI triples consists of localization loss and behavioral classification loss. Among them, the localization loss is calculated as in Eq. (\ref{eq:4}):

\begin{equation}
    \begin{aligned}
        \mathcal{L}_{loc}(c_i^h,c_i^o,a_{\sigma\left(i\right)})=
        &-log\frac{\text{exp}(\text{sim}(\text{FFN}_h(a_{\sigma(i)}),\mu_{c_i^h})/\tau)}{\sum_{i=1}^{K}{\text{exp}(\text{sim}(\text{FFN}_h(a_{\sigma(i)}),\mu_k)/\tau)}} \\
        &-log\frac{\text{exp}(\text{sim}(\text{FFN}_o(a_{\sigma(i)}),\mu_{c_i^0})/\tau)}{\sum_{i=1}^{K}{\text{exp}(\text{sim}(\text{FFN}_o(a_{\sigma(i)}),\mu_k)/\tau)}},
    \end{aligned}
    \label{eq:4}
\end{equation}

\noindent
where $\sigma\left(i\right)$ represents a set of matched pairs of human entities and object entities indexed $i$, $a_{\sigma\left(i\right)}$ is the behavioral prediction between the two entities, and $\tau$ is a temperature parameter controlling the smoothness of the loss function. The behavioral matching cost is calculated as $\mathcal{L}_{act}(a_i, {\hat{a}}_{\sigma(i)})=\text{BCELoss}(a_i, {\hat{a}}_{\sigma(i)})$ as in Eq. (\ref{eq:5}):

\begin{equation}
    \mathcal{L}_{act}(a_i, {\hat{a}}_{\sigma(i)})= -[a_i \cdot \log(a_{\sigma(i)}) + (1 - a_i) \cdot \log(1 - a_{\sigma(i)})].
    \label{eq:5}
\end{equation}

Subsequently, the Hungarian loss for all the previously identified matched pairs is computed according to the Eq. (\ref{eq:6}):

\begin{equation}
    \mathcal{L}_H=\sum _{i=1}^K[\mathcal{L}_{loc}(c_i^h, c_i^o,z_{\sigma(i)})+\mathcal{L}_{act}(a_i,a_{\sigma(i)})].
    \label{eq:6}
\end{equation}

\subsubsection{HOI-Caption generation decoder}
\label{subsec3.1.2}
The HOI-Caption generation decoder, which consists of the Image-HOI encoder and the Image-HOI-Caption decoder, aims to achieve efficient threat detection and generate interpretable text. Specifically, Image-HOI encoder processes images and assigns image embeddings and HOI guiding tags to a shared embedding space, facilitating multimodal feature integration through a cross-attention mechanism. This enhances the semantic expression of threat events in images. Subsequently, Image-HOI-Caption decoder generates explanatory text based on the encoded features. Interpretable text is used as a supervisory signal in the training phase to maximize the generation probability through autoregressive language modeling loss:

\begin{equation}
    \mathcal{L}_{LM}=-E_{X\to D}[CE(x,P(x))]=-E_{X\to D}[\sum_{i=1}^Nlog(P(x_i|x_{<i}))],
    \label{eq:7}
\end{equation}

\noindent
where $x_i$ denotes the $i^{\text{th}}$ token in the text and $N$ denotes the total number of text tokens. This mechanism ensures that the output text accurately portrays and clearly explains the threat event at the semantic level.

\subsubsection{Projector}
\label{subsubsec3.1.3}

In order to achieve effective alignment of visual features with encoded HOI-pairs, this article introduces a projector consisting of two layers of linear transformations and one layer of ReLU activation functions to map the two to a uniform embedding space for cross-modal feature fusion. It is imperative to note that the linear layer within this module remains non-frozen during the model fine-tuning phase.

\subsection{Training of Hoi2Threat}
\label{subsec3.2}

In this article, we employ well-taged fine-grained HOI-pairs for the purpose of supervised training of HOI encoder. In order to enhance the model's ability to model complex interactions in threat events, the model is further extended from person-to-person relationships to object-to-object interactions (indirect threat scenarios such as traffic accidents). This extension aims to improve the model's ability to understand and recognize diverse semantic interaction patterns.

In the training phase of the text generation, the Projector is trained using interpretive text and visual information, and some of the parameters of the HOI-Caption Generation decoder are fine-tuned to ensure the ability of the generated text to represent the input information. Furthermore, HOI-pairs are incorporated as explicit semantic guides to align the union of images and texts in the HOI representation space:

\begin{equation}
    \text{Image}_i \buildrel \text{Detect} \over \longrightarrow \text{Hoi}_i, \text{Text}_i \buildrel \text{Align}. \over \longrightarrow \text{Hoi}_i,
    \label{eq:8}
\end{equation}
\noindent
where $\text{Image}_i$ denotes the $i^{\text{th}}$ image input, $\buildrel \text{Detect} \over \longrightarrow$ denotes the HOI-pairs in the image extracted and encoded by the HOI encoder, $\text{Text}_i$ denotes the descriptive text corresponding to $\text{Image}_i$, and $\buildrel \text{Align} \over \longrightarrow$ denotes aligning the text with the HOI-pairs in the training data generation stage to ensure that the text content can completely cover each constituent element in the HOI-pairs. This explicit semantic guidance has two primary benefits. First, it improves the accuracy of text generation. Second, it enhances the model's ability to characterize the key elements and interactions of threat events. Consequently, it provides robust information support for subsequent risk decisions.

\section{Experiment}
\label{sec4}
In this section, we conducted extensive experiments to fully validate and demonstrate the superior performance of Hoi2Threat in threat detection and explanatory text generation tasks.

\subsection{Dataset}
\label{subsec4.1}
The training data for the TD-Hoi dataset was been derived from two sources: the largest Video Anomaly Detection (VAD) benchmark dataset currently available, XD-Violence, and the largest general-purpose dataset, COCO. The dataset captured a total of 8,282 images from Flickr, surveillance cameras, movies, in-car cameras, and games. It encompassed a wide range of imagery, including unusual and everyday scenes, as well as various threat image types, such as assaults, abduct, car accidents and more. Of these, 3,311 were classified as threat images and 4,971 as unusual images. To verify the excellent performance of Hoi2Threat on diverse datasets, we randomly selected video frame images from three datasets, Real Life Violence Situations \citep{9530829}, Large-scale Anomaly Detection \citep{wan2021anomalydetectionvideosequences}, and UCF-Crimes. Since this article focuses on scenarios that contain HOI, we exclude images that do not involve HOI scenarios to construct a test image set for performance evaluation. In light of these findings, we employ GPT-4.1 to generate HOI guiding tags and description text for each image in the test set. In order to ensure the reliability of the truth values of the test set, we manually verified and corrected the generated HOI guiding tags and descriptive text. The final stage of the experiment involved the evaluation of Hoi2Threat's performance using a set of 1,768 images.

\subsection{Metrics}
\label{subsec4.2}
In this section, we proposed an evaluation method, termed the Threat Image - based Text Generation Performance Benchmarking evaluation method, which was employed to assess the text generation performance of Image - based dialog models. In comparison with conventional text generation evaluation metrics, such as CIDEr \citep{Vedantam2015CIDEr} and BLEU \citep{Papineni2002BLEU}, this evaluation method demonstrated reduced susceptibility to the influence of tautological substitution or logical structure transformation. Consequently, it could provide a more reliable foundation for assessing text generation quality. To ensure objective and equitable evaluation outcomes, this article employed a LLM as the evaluation instrument. However, relevant studies had shown that LLMs exhibited a propensity to assign higher scores to their own 'low confusion' texts, thus significantly favoring responses that aligned with their own styles in the evaluation process. This phenomenon was referred to as Self-Preference Bias (SPB). In order to minimize the impact of SPB on the assessment results, this article abandoned the GPT-4.1 model and instead used the Deepseek-V3-0324 model to generate predicted scores on the following three key dimensions:

(1) Correctness of information: examine the consistency between the subject of the event involved in the generated text and the content of the image, focusing on assessing whether there is any misdetection or omission.

(2) Behavior mapping accuracy: measure whether the model can accurately map threat behaviors to the correct related entities, to avoid the focus of understanding the threat situation being shifted due to mapping errors.

(3) Threat detail orientation: with the help of the analysis of threat behaviors and their related entities in the image, to assess the information granularity of the generated text in describing the threat events, instead of staying in a generalized representation.

In the relationship-guided tag generation task, five evaluation metrics were used: $Precision$, $Recall$, $F1-\text{score}$, $Jaccard$ and $\text{Top}-K$ accuracy (k = 1, 3, or 5). These metrics were utilized to assess the tag generation capabilities of diverse models. $Precision$ is defined as the ratio of the number of correctly predicted tags to the total number of predicted tags in each sample. $Recall$ is defined as the ratio of the number of correctly predicted tags to the number of true-positive tags in each sample. The $F1-\text{score}$ is the harmonic mean of $Precision$ and $Recall$. The $Jaccard$ coefficient is the ratio of the size of the intersection of the predicted tags and the true tags to the size of their union in each sample. $\text{Top}-K$ accuracy is defined as the situation where at least one of the top K predicted tags is a true tag. The formula is defined as shown in Eq. (\ref{eq:9}):

\begin{equation}
    \begin{aligned}
        &Precision=\frac{1}{N}\sum_{i=1}^N\frac{TP_i}{TP_i+FP_i} \\
        &Recall=\frac{1}{N}\sum_{i=1}^N\frac{TP_i}{TP_i+FN_i} \\
        &F1-\text{score}=\frac{2*Precision*Recall}{Precision+Recall} \\
        &Jaccard=\frac{1}{N}\sum_{i=1}^N\frac{TP_i}{TP_i+FP_i+FN_i} \\
        &\text{Top}-K=\frac{m}{N},
    \end{aligned}
    \label{eq:9}
\end{equation}

\noindent
where $N$ is the total number of samples, $TP_i$ is the number of correctly predicted positive tags in the $i$ sample, $FP_i$ is the number of incorrectly predicted positive tags in the $i$ sample, $FN_i$ is the number of true positive tags in the $i$ sample that were not predicted, $m$ is the number of first $K$ predicted tags in which at least one of them is a true tag. With these multidimensional evaluation metrics, we are able to reflect more comprehensively the actual performance of the model in the tasks of HOI extraction and explanatory text generation.

\subsection{Implementation details}
\label{subsec4.3}
In regard to the training parameters, the AdamW optimizer is employed to train the model for a total of 10 epochs. The HOI encoder initial learning rate is set to 5e-6. The HOI-Caption generation decoder learning rate is 1e-4, the batch size is 36. The training and testing of this article is carried out using 4 $\times$ NVIDIA RTX A6000 GPUs.

\subsection{Main results}
\label{subsec4.4}
In order to rigorously evaluate the performance of Hoi2Threat in threat tasks in human violence scenarios, a thorough comparison is conducted with current state-of-the-art open source multimodal methods. The response results of each method were collected on a test set and submitted to Deepseek-V3-0324 for evaluation. The evaluation framework employs three key metrics to assess model performance: correctness of information (CoI), behavioral mapping accuracy (BMA), and threat detail orientation (TDO). Each metric is assigned a score ranging from 1 to 10, with higher values denoting enhanced performance of the model in that specific metric. The comprehensive evaluation results of all models are presented in Table \ref{tab1}.

\begin{table}[htpb]
    \centering
    \scriptsize
    \caption{A comparison with state-of-the-art threat detection methods.}
    \begin{tabularx}{\textwidth}{|c|c|*{9}{c|}} 
        \hline
        \multirow{2}{*}{Models} & \multirow{2}{*}{\makecell{Para-\\ meters}} & \multicolumn{3}{|c|}{\makecell{Large-scale\\ Anomaly Detection}} & \multicolumn{3}{|c|}{\makecell{RealLife Vio- \\ lence Situations}} & \multicolumn{3}{|c|}{UCF-Crimes} \\
        &  & CoI & BMA & TDO & CoI & BMA & TDO & CoI & BMA & TDO \\\hline
        LLama3.2-V& 11B& 3.19& 2.98& 2.89& 3.27& 3.39& 3.28& 3.04& 2.89& 2.81\\
        GLM4& 9B& 1.39& 1.32& 1.32& 1.26& 1.33& 1.26& 1.30& 1.24& 1.23\\
        LLava-LLama3& 8B& 2.83& 2.68& 2.62& 2.33& 2.40& 2.34& 2.76& 2.64& 2.57\\
        MiniCPM-V& 8B& 3.49& 3.32& 3.2& 3.49& 3.59& 3.50& 3.27& 3.13& 3.06\\
        Gemma3& 4B& 5.13& 5.15& 4.86& 4.33& 4.25& 4.43& 4.42& 4.41& 3.94\\
        MoonDream& 1.8B& 3.52& 3.36& 3.28& 3.52& 3.63& 3.54& 3.49& 3.34& 3.26\\
        Tag2Text& 0.37B& 2.19& 2.08& 2.02& 1.70& 1.75& 1.72& 1.99& 1.90& 1.86\\
        Ours& \textbf{0.35B}& \textbf{5.68}& \textbf{5.43}& \textbf{4.78}& \textbf{4.53}& \textbf{4.66}& \textbf{4.58}& \textbf{4.60}& \textbf{4.69}& \textbf{4.27}\\ \hline
    \end{tabularx}
    \label{tab1}
\end{table}

The experimental results demonstrate that Hoi2Threat significantly outperforms existing methods in CoI, BMA, and TDO on multiple benchmark datasets, thereby substantiating its advancement in the threat event detection task. With respect to the concept of key objects identified (CoI), Hoi2Threat can reliably capture targets while effectively mitigating misjudgment errors of regarding irrelevant objects as subjects or failing to identify significant targets. In terms of BMA metrics, Hoi2Threat can accurately establish the correspondence between threat behaviors and related entities and effectively reduce semantic mapping deviations. This ensures that the model’s understanding of threat scenarios highly aligns with the semantics of actual images. In terms of TDO metrics, Hoi2Threat generates more detailed descriptions of threat events, providing informative and specific representations and avoiding over-generalizations. Such detailed descriptions are particularly important in the analysis of semantically complex human violence scenarios. Furthermore, in contrast to other generative models, the BERT-based architecture of Hoi2Threat has a considerably reduced number of parameters, requiring fewer computational resources during operation. This effectively mitigates the computational burden of downstream applications and ensures their efficient operation in resource-limited environments.

\begin{figure}[htbp]
\centering
\includegraphics[width=\textwidth]{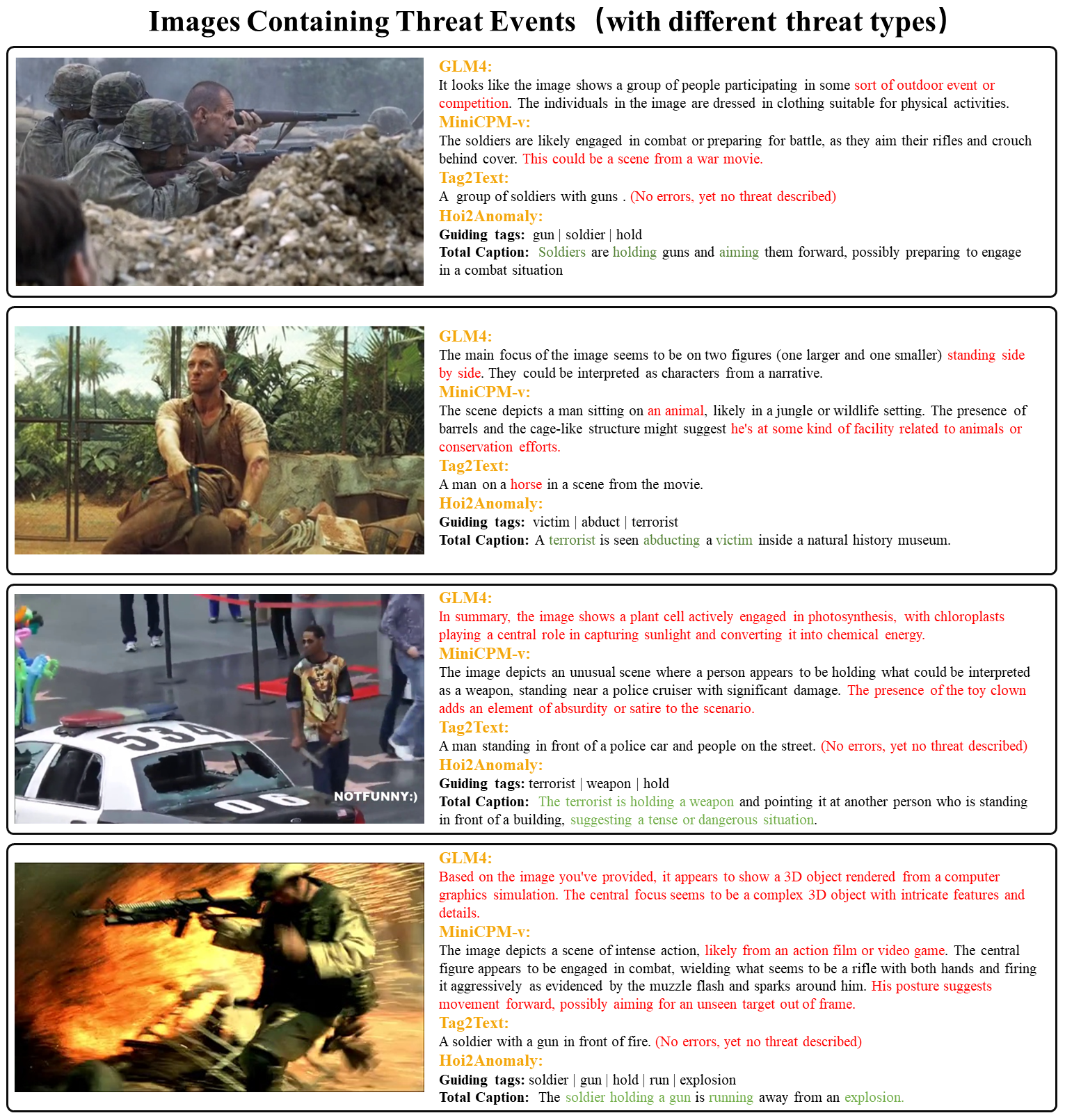}
\caption{Qualitative Results. We compare the performance of Hoi2Threat, GLM4, MiniCPM-V and Tag2Text. To visualize the model performance, correct interpretations are marked in green and incorrect interpretations are marked in red.}\label{fig4}
\end{figure}

To further validate the performance of the model, in this article four images of different types of threats were selected, as shown in Fig. \ref{fig4}. A unified cue was used to obtain the responses of GLM4 \citep{glm2024chatglm}, MiniCPM-V \citep{yao2024minicpm}, Tag2Text, and our proposed Hoi2Threat model, respectively. The experimental results demonstrated that GLM4 and Tag2Text solely offer more generalized image descriptions and are ineffective in extracting threat features. Although MiniCPM-V exhibits some threat recognition capability, it frequently misjudged, and the generated content often deviated from the image semantics. In contrast, Hoi2Threat not only accurately identifies threats but also provided detailed descriptions of the “content”, “behavior”, and “participants” of an event. The Hoi2Threat system’s incorporation of an explicit HOI tag guidance mechanism facilitates the accurate identification of the core elements of the event and the participating subjects. This mechanism enables the generation of structured explanatory text, which presents the semantic content of the threat situation in a comprehensible manner. Consequently, Hoi2Threat provides substantial support for subsequent analysis and decision-making processes.

\subsection{Analytical results}
\label{subsec4.5}
The experiment was meticulously designed to evaluate three critical components: the HOI encoder, the tag guidance generation mechanism, and the joint encoder introducing location coding. These components were evaluated across multiple dimensions to ascertain their effectiveness in capturing and expressing the features of the subject’s events. The second objective was to conduct tag guidance generation mechanism ablation experiments, with the aim of validating the contribution of tags in guiding the model to understand the semantics of the images and to generate accurate text. The third objective was to perform positional coding ablation experiments, with the aim of evaluating its role in assisting the model to focus on key regions and accurate semantic understanding. The importance of each module in threat target identification, behavior mapping, and detail portrayal is quantified by comparing the performance of the full model with each ablation variant on a multi-scenario threat event benchmark set.

\subsubsection{HOI encoder generative capability ablation experiments}
\label{subsubsec4.5.1}

In order to validate the key role of HOI encoder in threat event detection, we evaluated the performance of Tag2Text, RAM, RAM++, and our proposed method using five evaluation metrics: $Precision$, $Recall$, $F1-\text{score}$, $Jaccard$, and $\text{Top}-K$ accuracy (k = 1/3/5). The results are shown in Table \ref{tab2}.

\begin{table}[htpb]
    \centering
    \caption{HOI tags extraction encoder generative capability ablation experiments.}
    \scriptsize
    \begin{tabular}{|c|c|c|c|c|c|}
    \hline
    Models & Precision/Recall/F1-score & Jaccard & Top-1 & Top-3 & Top-5 \\\hline
    Tag2Text& 0.40/0.19/0.24 & 0.17 & 0.05 & 0.16 & 0.30 \\
    RAM& 0.43/0.20/0.26 & 0.18 & 0.17 & 0.32 & 0.36 \\
    RAM++& 0.44/0.19/0.26 & 0.18 & 0.19 & 0.32 & 0.36 \\
    Ours& \textbf{0.45/0.39/0.41} & \textbf{0.33} & \textbf{0.42} & \textbf{0.66} & \textbf{0.69} \\
    \hline
    \end{tabular}%
    \label{tab2}
\end{table}

As was demonstrated in Table \ref{tab2}, the proposed method exhibits a high degree of efficacy in terms of the $F1-\text{score}$ metric. It is noteworthy that elevated $F1-\text{score}$ are indicative of a superior capacity to comprehensively capture pertinent information within an image. The enhanced scores attained on the $Jaccard$ metrics signify a substantial increase in the overlap between the predicted outcomes and the true set, thereby suggesting that the model demonstrates a propensity to predict the key tags in an event. Furthermore, the consistent performance on the $\text{Top}-K$ metric serves to substantiate the efficacy of the model in tag ranking and key tag identification. The aforementioned results unequivocally demonstrate that the introduced HOI encoder is capable of generating high-quality and low-bias tags of threat behaviors and related entities, providing precise semantic guiding for the tag-guided generation mechanism, and thus significantly improving the model’s threat detection performance in complex violence scenarios.

\subsubsection{Tag-guided generation mechanisms and position embedding ablation experiments}
\label{subsubsec4.5.2}
In order to verify the effectiveness of the label guidance generation mechanism and position coding in image understanding guidance, we designed an experimental study with three models: no label guidance generation (without hoi tag), no position embedding (without pos), and the full model. In order to validate the model performance with greater rigor, a quantitative comparative analysis of the models at different training stages on the test set was conducted using Deepseek-V3-0324. The primary objective of this study is to provide a comparative analysis of the model's performance before and after the integration of tag-guided generation and position embedding. The results of this analysis are presented in Table \ref{tab3}. 

\begin{table}[htpb]
    \centering
    \scriptsize
        \caption{Ablation experiments of tag-guided generation mechanisms and position embedding.}
        \begin{tabular}{|c|*{9}{c|}} 
        \hline
        \multirow{2}{*}{Models} & \multicolumn{3}{|c|}{\makecell{Large-scale\\ Anomaly Detection}} & \multicolumn{3}{|c|}{\makecell{RealLife Vio- \\ lence Situations}} & \multicolumn{3}{|c|}{UCF-Crimes} \\
         & CoI & BMA & TDO & CoI & BMA & TDO & CoI & BMA & TDO \\
         \hline
        Ours (without hoi tag)&1.70 &1.64 &1.56 &1.23 &1.22 &1.34 &1.57 &1.54 &1.63
        \\
        Ours (without pos)&4.50 &4.62 &4.07 &4.05 &4.22 &4.05 &4.09 &4.24 &3.83
        \\
        Ours& \textbf{5.68}& \textbf{5.43}& \textbf{4.78}& \textbf{4.53}& \textbf{4.66}& \textbf{4.58}& \textbf{4.60}& \textbf{4.69}& \textbf{4.27}
        \\
        \hline
    \end{tabular}
    \label{tab3}
\end{table}

The findings indicate that the model's performance in critical evaluation metrics, including CoI, BMA and TDO, has undergone a substantial enhancement, with an average improvement of up to 160\% following the implementation of label-guided generation. This substantial enhancement in performance suggests that the label-guided generation mechanism plays a pivotal role in enhancing the model's capacity to discern threat details during the inference phase. This mechanism has the potential to assist the model in more accurately identifying the core targets and behaviors of an event, thereby significantly optimizing the model's overall performance in the image threat detection task. Furthermore, the implementation of location coding plays a pivotal role in the cross-attention embedding process, effectively guiding the model to prioritize key regions within the image, particularly the region where the threat event transpires. This enhancement in turn leads to an augmentation in the model's capacity to interpret the content of the images. The integration of positional coding has been instrumental in enhancing the model's capacity to generate more precise textual responses, accompanied by a substantial reduction in illusions. Consequently, the quality and reliability of the output results have undergone a notable enhancement.

In summary, Hoi2Threat is capable of providing precise descriptions of threat events in images and effectively suppressing hallucinations by utilizing cueing information regarding threat behaviors and related entities in images. When combined with the threat label-based cue text generation mechanism, the model enhanced its reliability and accuracy in complex image understanding tasks, highlighting its strong applicability in diverse scenarios. In the context of high-risk mission deployments, this ability enables Hoi2Threat to reduce the risk of information errors, providing strong technical support for ensuring the smooth progress of the tasks.

\section{Conclusion}
\label{sec5}
In this article, we proposed an image threat detection and interpretable text generation method for human violence scenarios, Hoi2Threat. This method uses fine-grained HOI-pairs as label guidance, fuses image features and interaction relations through cross-attention, and combines with location coding to realize accurate mapping of threat behaviors and entities. This significantly improves the detection accuracy and semantic consistency of the model in multi-entity complex scenarios. Meanwhile, ensure that the generated text responses not only accurately reflect the threat elements but also have a clear reasoning context. A substantial body of comparative experiments and ablation analyzes demonstrates the superiority of Hoi2Threat in key metrics such as CoI, BMA, TDO, and others. Additionally, its proficiency in detail portrayal and text interpretability is noteworthy. Hoi2Threat, in conjunction with its constructed TD-Hoi dataset, is poised to serve as a valuable resource for the research community, particularly in the realm of image threat detection. This will effectively stimulate the advancement of related technologies in the domain of public security.

\section*{Future work}
Image-level threat detection constitutes a fundamental component of video analytics, with its performance exerting a direct influence on the efficacy of video-level detection. Due to the limitation of article length, the present article principally focuses on image-level threat detection, yet it still exhibits good real-time video scalability. In the future, we intend to extend the current threat detection methods to the real-time video level through the implementation of keyframe selection and timing modeling techniques.

\section*{CRediT authorship contribution statement}
Yuhan Wang: Conceptualization, Methodology, Data curation, Validation, Writing–original draft, Writing–review \& editing. Cheng Liu: Conceptualization, Resources, Data curation, Writing–review \& editing. Daou Zhang: Resources, Data curation. Zihan Zhao: Data curation, Writing–review \& editing. Jinyang Chen: Data curation. Purui Dong: Data curation. Ziru Wang: Data curation. Zuyuan Yu Chen: Data curation. Weichao Wu: Conceptualization, Funding acquisition, Resources, Project administration, Writing–review \& editing, Supervision.

\section*{Declaration of competing interest}
The authors declare that they have no known competing financial interests or personal relationships that could have appeared to influence the work reported in this article. 
All our models and data are used for research purposes only to avoid potential negative social impacts.

\section*{Declaration of generative AI and AI-assisted technologies in the writing process}
During the preparation of this work the authors used DeepL’s DeepL Write and GPT-4.1 to polish the language of the article.

\section*{Data availability}
Data and code will be made available at \url{https://github.com/XsjFyZs7/Hoi2Threat}

 



\bibliography{reference} 

\end{document}